\def\year{2021}\relax
\documentclass[letterpaper]{article} 
\usepackage{aaai21}  
\usepackage{times}  
\usepackage{helvet} 
\usepackage{courier}  
\usepackage[hyphens]{url}  
\usepackage{graphicx} 
\usepackage{natbib}  
\usepackage{caption} 
\usepackage{researchpack}
\usepackage{xcolor}
\usepackage{xspace}
\usepackage{hyperref}
\usepackage{subfig}
\usepackage[super]{nth}
\usepackage{wrapfig}
\usepackage{booktabs}
\hypersetup{hidelinks}
\usepackage{cite}
\newcommand{\teodora}[1]{}
\newcommand{\mohit}[1]{}
\newcommand{\stefano}[1]{}

\title{Modeling High-order Interactions across Multi-interests for Micro-video Recommendation (Student Abstract) }
\author{
    Dong Yao\textsuperscript{\rm 1},
    Shengyu Zhang\textsuperscript{\rm 1},
    Zhou Zhao\textsuperscript{\rm 1},  \\
    Wenyan Fan\textsuperscript{\rm 1},
    Jieming Zhu\textsuperscript{\rm 2},
    Xiuqiang He\textsuperscript{\rm 2},
    Fei Wu\textsuperscript{\rm 1}
    \\
}
\affiliations{
	\textsuperscript{\rm 1}Zhejiang University\\
	\textsuperscript{\rm 2}Huawei\\
	\{yaodongai, sy\_zhang, zhaozhou, wenyan.17, wufei\}@zju.edu.cn\\
		
		jmzhu@ieee.org, hexiuqiang1@huawei.com\\
	}

\begin{document}
	
	\maketitle
	
	\begin{abstract}
		Personalized recommendation system has become pervasive in various video platform. Many effective methods have been proposed, but most of them didn't capture the user's multi-level interest trait and dependencies between their viewed micro-videos well. To solve these problems, we propose a Self-over-Co Attention module to enhance user's interest representation. In particular, we first use co-attention to model correlation patterns across different levels and then use self-attention to model correlation patterns within a specific level. Experimental results on filtered public datasets verify that our presented module is useful.  
	\end{abstract}

	\section{Introduction}
	With the emergence of various micro-videos platforms, the amount of micro-videos and users are growing exponentially~\cite{titling}. Simultaneously, it also becomes increasingly difficult for users to watch micro-videos that they are interested in~\cite{poet}. Given this, an intelligent personalized recommendation system is crucial. However, building a personalized recommendation system is a big challenge, owing to the following two reasons:1) $\textbf{Multi-level interest}$. There are different interaction behavior between users and micro-videos. In our dataset, three interaction types are included: ``click", ``like", and ``follow", representing user's different levels of preference. 2) $\textbf{Correlation and dependencies}$. Various micro-videos interacted by one user may have inherent correlation from each, and the pattern of such correlation can vary within and across levels of interest.
	
	In recent years, many effective models have been applied in the personalized recommendation system, such as collaborative filtering based models~\cite{a}, content-based systems~\cite{content}, hybrid methods~\cite{item} and sequential recommendation~\cite{future}. Despite the promising performance achieved by such methods, most of them neglect the nature of the user's multi-level interests. To this end, some researchers propose to model the user's multi-level interest and construct level-specific embedding in a pre-processed manner~\cite{routing}. However, they don't consider the correlation of micro-videos. To explicitly capture such correlation, we propose the Self-over-Co Attention module, dubbed as \emph{SoC Attention}. The overall architecture is displayed in Figure 1. Specifically, SoC Attention first models the correlation patterns across levels using co-attention and investigates the level-specific correlation using the self-attention mechanism. Compared with existing methods that directly learn the level related representation by back-propagation such as ALPINE ~\cite{routing}, SoC Attention can quickly learn the level related representation of users beyond the training set without retraining. We construct a model named SCAA based on the proposed SoC Attention module and the ALPINE-Base model which is a modified version of ALPINE without its original multi-level interest module.

	\section{Our Method}
	
	In this section, we will illustrate the two building blocks of SoC Attention in detail:	
	\subsection{Co-attention Layer}
	Expressly, for a user, we set a set of micro-videos' features as $\mathbf{U}_{l}$, in which the user likes every video. Analogously, we develop a set of micro-videos' features as $\mathbf{U}_{f}$, in which the user follows every video. Firstly, we feed $\mathbf{U}_{l}$ into three full connected layers that each of them has different parameters,
	$$
	\mathbf{Q}_{l}=\mathbf{W}_{ql}\mathbf{U}_{l},\mathbf{K}_{l}=\left(\mathbf{W}_{kl}\mathbf{U}_{l}\right)^{T},\mathbf{V}_{l}=\mathbf{W}_{vl}\mathbf{U}_{l}
	$$
	Where $\mathbf{Q}_{l}$, $\mathbf{K}_{l}$, $\mathbf{V}_{l}$ are query, key, value matrix for liked micro-videos. Likewise, We feed $\mathbf{U}_{f}$ into three various full connected layers,
	$$
	\mathbf{Q}_{f}=\mathbf{W}_{qf}\mathbf{U}_{f}, \mathbf{K}_{f}=\left(\mathbf{W}_{kf}\mathbf{U}_{f}\right)^{T}, \mathbf{V}_{f}=\mathbf{W}_{vf}\mathbf{U}_{f}
	$$
	where $\mathbf{U}_{l} \in \mathbb{R}^{m \times d}$, $\mathbf{U}_{f} \in \mathbb{R}^{n \times d}$, and $\mathbf{Q}_{f}$, $\mathbf{K}_{f}$, $\mathbf{V}_{f}$ are query, key, value matrix for followed micro-videos. We set $\mathbf{softmax}$ funuction as $g(\cdot)$. Then, the enhanced ``like" and ``follow" visual features $\mathbf{U}_{l}^{e}$ are computed as follows, 
	$$
	\mathbf{U}_{l}^{e}=g(\mathbf{Q}_{l}\mathbf{K}_{f})\mathbf{V}_{f}+\mathbf{U}_{l}, \mathbf{U}_{f}^{e}=g(\mathbf{Q}_{f}\mathbf{K}_{l})\mathbf{V}_{l}+\mathbf{U}_{f}
	$$    
	\subsection{Self-attention Layer}
	Thereafter, we feed $\mathbf{U}_{l}^{e}$ into a group of full-connected layers, 
	$$
	\mathbf{Q}_{l}^{e}=\mathbf{W}_{ql}^{e}\mathbf{U}_{l}^{e}, \mathbf{K}_{l}^{e}=\left(\mathbf{W}_{kl}^{e}\mathbf{U}_{l}^{e}\right)^{T}, \mathbf{V}_{l}^{e}=\mathbf{W}_{vl}^{e}\mathbf{U}_{l}^{e}
	$$
	\begin{figure}[htb]
		\centering
		\includegraphics[width=\columnwidth]{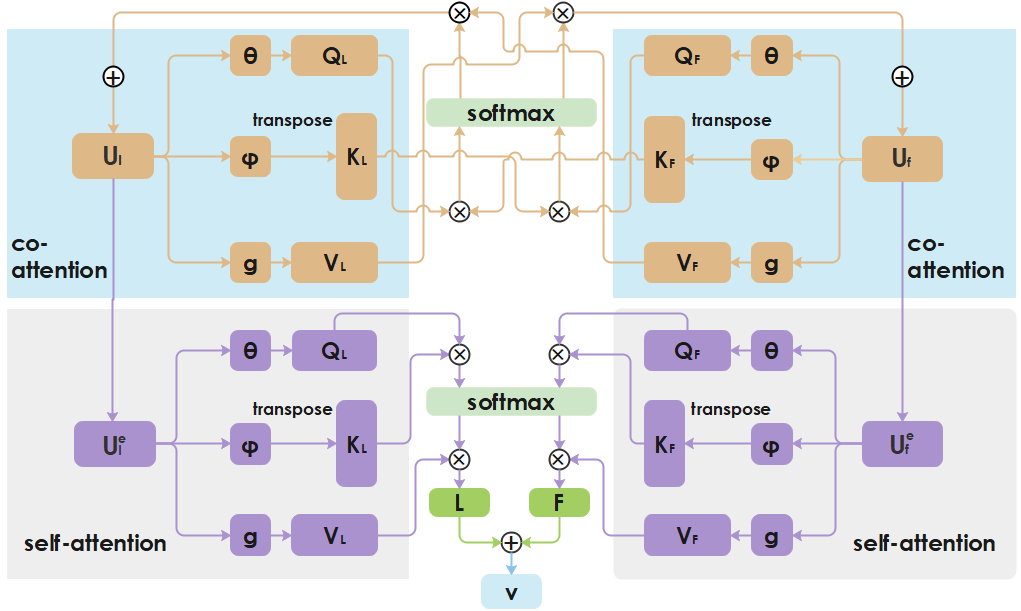}
		\caption{SoC Attention module}
		
	\end{figure}
	Where $\mathbf{Q}_{l}^{e}$, $\mathbf{K}_{l}^{e}$, $\mathbf{V}_{l}^{e}$ denote enhanced query, key, value matrix for liked micro-videos. For the $\mathbf{U}_{f}^{e}$, 
	$$
	\mathbf{Q}_{f}^{e}=\mathbf{W}_{qf}^{e}\mathbf{U}_{f}^{e}, \mathbf{K}_{f}^{e}=\left(\mathbf{W}_{kf}^{e}\mathbf{U}_{f}^{e}\right)^{T}, \mathbf{V}_{f}^{e}=\mathbf{W}_{vf}^{e}\mathbf{U}_{f}^{e}
	$$
	where $\mathbf{U}_{l}^{e} \in \mathbb{R}^{m \times d}$, $\mathbf{U}_{f}^{e} \in \mathbb{R}^{n \times d}$, and $\mathbf{Q}_{l}^{e}$, $\mathbf{K}_{l}^{e}$, $\mathbf{V}_{l}^{e}$ denote enhanced query, key, value matrix for followed micro-videos. Then, we get the ``like" interest matrix $\mathbf{L}=\left[\mathbf{l}_{1},\mathbf{l}_{2},...,\mathbf{l}_{m}\right]^{T}$ and the ``follow" interest matrix $\mathbf{F}=\left[\mathbf{f}_{1},\mathbf{f}_{2},...,\mathbf{f}_{n}\right]^{T}$ shown in the following equations, 
	$$
	\mathbf{L}=g(\mathbf{Q}_{l}\mathbf{K}_{l})\mathbf{V}_{l}, 
	\mathbf{F}=g(\mathbf{Q}_{f}\mathbf{K}_{f})\mathbf{V}_{f}
	$$ 
	Eventually, we get final improved interest representation $\mathbf{v}$,
	$$
	\mathbf{l}=\frac{1}{m}\sum_{i=1}^{m}\mathbf{l}_{i},
	\mathbf{f}=\frac{1}{n}\sum_{i=1}^{n}\mathbf{f}_{i},
	\mathbf{v}=\frac{m}{m+n}\mathbf{l}+\frac{n}{m+n}\mathbf{f}
	$$ 
	\section{Experiment}
	
	\subsection{Dataset and Evaluation Metric}
	We use the dataset released by the Kuaishou Competition\footnote{https://anonymous1240.wixsite.com/alpine}. Then, we divide it into two parts: \romannumeral1) 80\% of the data as the training set; \romannumeral2) 20\% of the data as the testing set. We select Area Under Curve(AUC), R@50, P@50, F@50 as evaluation metric. For a recommendation list computed based on the click probability, R@50, P@50, separately means the recall value, precision of the top50 items, and F@50 is the harmonic average of corrsponding precision and recall. 
	\subsection{Comparative Experiment and Ablation Study}
	To testtify the effectiveness of SCAA, we compare it with: 
	\romannumeral1) ATRank~\cite{atrank} \romannumeral2) NCF~\cite{ncf} \romannumeral3) THACIL~\cite{thacil}. The results of the 
	aforementioned models are shown in Table 1. 
	It shows that SCAA doesn't achieve better 
	\begin{table}[htb]
		\centering
		\setlength{\tabcolsep}{1.3mm}{
			\linespread{1.1}\selectfont
			\begin{tabular}{cccccc}  \toprule
				Methods  &ATRank &NCF &THACIL &ALPINE &SCAA\\  \hline
				AUC &0.722 &0.724 &0.727 &\textbf{0.737} &\textbf{0.737} \\
				\hline
		\end{tabular}}
		\caption{Performance of SCAA, ALPINE and other models 
		} 
	\end{table}
	performance compared with ALPINE. Actually, most of the users in the dataset don't have ``like" and ``follow" interaction behavior. Obviously, SoC Attention doesn't work on these users. Therefore, we filter these users and retrain models. The result is showed in Table 2. It is worth mentioning that SCAA without SoC Attention is our baseline model. The improvement of SCAA over 
	ALPINE is 2.3\%, reflecting that the effectiveness of SoC Attention on gaining user's interest representation is distinct.     
	\begin{table}[t]

		\centering
		\setlength{\tabcolsep}{3.3mm}{
			
			\begin{tabular}{l l l l l}  \toprule
				Methods 
				&AUC &R@50 &P@50 &F@50\\  \hline
				
				ALPINE 
				&0.696 &0.383 &0.355 &0.368\\
				
				SCAA\_cs
				& 0.701 &0.383 &0.355 &0.368 \\
				SCAA\_s
				& 0.705 &0.385 &0.359 &0.371\\
				SCAA
				&\textbf{0.712} &\textbf{0.390} &\textbf{0.364} &\textbf{0.376} \\  
				\hline
		\end{tabular}}
	\caption{Performance of the original ALPINE, SCAA and several variants on the filtered dataset }
	\end{table}
	We design ablation experiments to study how each component in SoC Attention contributes to the final performance: $\textbf{SCAA\_cs:}$ We removed the Co-attention part and Self-attention part from SCAA; $\textbf{SCAA\_s:}$ We eliminated the Self-attention part from SCAA. The results are demonstrated in Table 2. By analyzing Table 2, we can conclude:	
	
    Compared with SCAA, SCAA\_s decreases by 1\% in terms of AUC and other indicators all decline. This revealed that the Self-attention part is useful in capturing the dependency of micro-videos of the same interaction behavior.
		
    It can be seen that SCAA\_s surpasses SCAA\_cs by 0.6\% in terms of AUC. Other indicators have a different degree of improvement. This indicates the Co-attention part's effectiveness in catching the correlation of micro-videos of the various interaction behavior.

\bibliography{Formatting-Instructions-LaTeX-2021}
\bibliographystyle{aaai}
	
\end{document}